\documentclass{iosart2c}

\usepackage{url}
\usepackage{enumitem}
\usepackage{mathptmx}
\usepackage{soul}\setuldepth{article}
\usepackage{graphicx}
\def\hb{\hbox to 11.5 cm{}}
\def\ONT{\textsc{OASIS}}

\def\OLDFW{\textsc{Prof-Onto}}
\def\FW{\textsc{Clara}}

\begin{document}

\pagestyle{headings}
\begin{frontmatter}              

\title{The Ontology for Agents, Systems and Integration of Services: OASIS version 2\thanks{This article is an extended revision of \textquotedblleft The Ontology for Agents, Systems and Integration of Services: recent advancements of OASIS", Proceedings of WOA 2022: 23rd Workshop From Objects to Agents, September 1--2, Genova, Italy. CEUR Workshop Proceedings, ISSN 1613-0073, Vol. 3261, pp.176--193. }}


\author[A]{\fnms{Giampaolo} \snm{Bella},}
\author[A]{\fnms{Domenico} \snm{Cantone},}
\author[A]{\fnms{Carmelo Fabio} \snm{Longo},}
\author[A]{\fnms{Marianna} \snm{Nicolosi Asmundo}}
and
\author[A]{\fnms{Daniele Francesco} \snm{Santamaria} \thanks{Corresponding Author: Daniele Francesco Santamaria, daniele.santamaria@unict.it}}

\address[A]{Department of Mathematics and Computer Science, University of Catania, Viale Andrea Doria 6 - 95125 - Catania, Italy}

\begin{abstract}
 Semantic representation is a key enabler for several application domains, and the multi-agent systems realm makes no exception. Among the methods for semantically representing agents, one has been essentially achieved by taking a behaviouristic vision, through which one can describe how they  operate and engage with their peers. The approach essentially aims at defining the operational capabilities of agents through the mental states related with the achievement of tasks.  The \ONT{} ontology --- An Ontology for Agent, Systems, and Integration of Services, presented in 2019 --- pursues the behaviouristic approach to deliver a  semantic representation system and a communication protocol for agents and their commitments. This paper reports on the main modeling choices concerning the representation of agents in \ONT{} 2, the latest major upgrade of \ONT, and the achievement reached by the ontology since it was first introduced, in particular in the context of ontologies for blockchains. 
\end{abstract}

\end{frontmatter}

\section{Introduction}
The gist of the \emph{Semantic Web} \cite{bernerslee2001semantic} is to achieve the full interoperability of software applications by means of reusing and sharing of data in standard formats across systems, enterprises, and community boundaries. In the Semantic Web vision of the Internet,  software agents are enabled to query and manipulate information on  behalf of human agents by means of machine-readable data. Such data carry explicit meaning, expressed with formally defined semantics through suitable representation languages that support the automatic processing of information. For this purpose, the \emph{Word Wide Web Consortium} (W3C) conceived the \emph{Web Ontology Language} (\textbf{OWL})~\cite{OWL2}, a family of knowledge representation languages relying on \textit{Description Logics}~\cite{baader}, as a standard for representing Semantic Web ontologies.\footnote{An ontology is  a set of representational primitives, typically classes, attributes and relationships, defined to formally model a domain of knowledge or discourse.}  
The main advantage of Semantic Web is to enable users and applications to access data-oriented web content: users and applications become able to discover and invoke web resources in a completely automatic way.
To capture the complex interactions of the participants in those environments  where people and organizations dynamically interact with autonomous systems, it is convenient to identify the proactive stakeholders as autonomous agents \cite{jennings98}. Agents are defined as entities whose state is perceived as a set of mental components such as beliefs, capabilities, choices, and commitments \cite{Shoham1992AgentOP}. This definition gives rise to the foundations for the \emph{Agent Oriented Programming}  (AOP) paradigm, which is most often motivated by the need for open architectures that continuously evolve to adapt to external changes in a robust, autonomous, and proactive way. Coherently with this vision, defining suitable representation mechanisms for agents is still challenging in the large realm of \emph{Knowledge Representation}.

The authors' choice for the semantic description of agents focuses on the agents' mental states related with their capabilities of reaching an achievement. Then, the behaviouristic approach, inspired by the \emph{Tropos} methodology \cite{tropos}, provides a practical and complete operational semantics. Agent behaviours are decomposed in the atomic and essential mental states of the agent, in such a way that each single task that the agent seeks to accomplish is completely described. The \emph{Ontology for Agents, Systems and Integration of Services} \cite{woa2019, woa2022}, \ONT{} in short, applies the behaviouristic approach to deliver a representation system and a communication protocol for agents and their commitments. Agents are enabled to manifest the set of operations that they are able to perform and the type of data required to execute them. Each output is also identified in a clear, human-understandable and unambiguous way, in order to abstract from the implementation details and transparently automate the task of agent discovery. Agents may then join a collaborative environment in a \emph{plug-and-play} way, since  third-party interventions are no longer required, thereby profiting from the Semantic Web features.  
By means of Semantic Web technologies and of automated reasoners, data provides machine-understandable information that can be processed, integrated and exchanged by any type of agent, at a higher level; data consistency can be easily verified, and information can be inferred and retrieved through the defined knowledge base. This leads to the realization of Smart Agents, IA-driven agents that consume information and act with minimal human intervention. Where required, agents are able to communicate with humans by using common natural interaction practises, like vocal commands, sentences transmitted via texts or gestures.

This article revises a previous workshop paper~\cite{woa2022} by reporting on the full modeling choices behind the latest version of \ONT{}, named \ONT{} 2. In brief, the new ontology reviews the representation of agents and their commitments and also introduces the constitutional elements that make it a valid means for defining a \emph{Semantic Blockchain}.
The full \ONT{} 2 is available online~\cite{oasis2}.

The paper is organized as follows. Section \ref{sec:ontology} introduces the main features of \ONT{} 2 and how it models agent behaviours and commitments. Section \ref{sec:contrib} reports on the contributions given by \ONT{}, depicting the main differences between \ONT{} and \ONT{} 2. Section \ref{sec:related} presents  related works discussing similarities and differences with \ONT{} 2, while Section \ref{ref:conclusions} concludes the paper with future perspectives.

\section{The \ONT{} 2 ontology} \label{sec:ontology}

This section  is devoted to the \ONT{} 2 modeling approach for representing agents and their commitments.  \ONT{} 2 is a foundational OWL 2 ontology that leverages the behaviouristic approach derived from the \emph{Theory of Agents} and the related mentalistic notions to represent multi-agent systems. The approach aims at describing how agents behave with respect to the environment by representing their essential mental states, namely  \emph{behaviours}, \emph{goals}, and \emph{tasks}.  Behaviours represent the mental state of the agent associated with its capability of activating, modifying the environment or doing something; goals describe mental attitudes representing preferred progressions of a particular system that the agent has chosen to put effort into bringing about \cite{riemsdijk08}; tasks depict how such progressions are carried on and describe atomic operations that agents may perform.

Such behaviouristic approach is an effective way for semantically describing agents by characterizing their capabilities. Agents are enabled to report the set of activities that they are able to perform, the types of data required to execute those activities as well as the expected output through the description of their behaviours. Agents' implementation  details are abstracted away so as to make the discovery of agents transparent, automatic, and independent of the underlying technical infrastructure. As a consequence, agent commitments are clearly described and the entire evolution of the environment is unambiguously represented, searchable, and accessible:  agents may join a collaborative environment in a plug-and-play fashion, as there is no more need of third-party interventions. The behaviouristic approach is exploited by \ONT{} 2 to characterize agents in terms of the actions they are able to perform, including purposes, goals, responsibilities, information about the world they observe and maintain, and their internal and external interactions. Finally, \ONT{} 2 models the executions and assignments of tasks, restrictions, and constraints used to establish agent responsibilities and authorizations. 

In \ONT{} 2, agents and their interactions are represented by carrying out three main steps, namely 
a) an optional step that consists in modeling general abstract behaviours (called \emph{templates}) from which concrete agent behaviours are drawn, b) modeling concrete agent behaviours, possibly, drawn by agent templates, and c) modeling actions and associating them with the corresponding behaviours.

The first step, not mandatory,  consists in defining the agent's behaviour template, namely a higher-level description of the behaviour of abstract agents that can be implemented to define more specific and concrete behaviours of real agents; for example, a template may be designed to describe which features a trader agent should own. Moreover, templates are useful to guide developers in the definition of the behaviours of their specific agents. To describe  abstract agent's capabilities to perform actions, an agent template comprises three main elements, namely behaviour, goal, and task.  In fact, the core of the agent representation mechanism in \ONT{} 2 revolves around  the description of atomic operations introduced by the definition of tasks, i.e.,  the most simple (atomic) operations that agents are able to actually perform including, possibly, input and output parameters required to accomplish them.

The second step consists in representing concrete agent behaviours either by specifying a shared template or by defining it from scratch. In both cases, concrete behaviours are modeled analogously to those of templates, where the models of ideal features are replaced with actual characteristics. Behaviours drawn by shared templates are clearly associated with them in order to depict the behaviour inheritance relationship.

In the last step, actions performed by agents are described as direct consequences of some behaviours, and are associated with the behaviours of the agent that performed them. To describe such an association, \ONT{} 2 introduces \emph{plan executions}. Plan executions describe the actions performed by an agent and associated with one of its behaviours. Associations are carried out by connecting the description of the performed action to the behaviour from which the action has been drawn:  actions are hence described by suitable graphs that retrace the model of the agent's behaviour. 
Additionally, plans can be either submitted to  agents as request of performing some actions or they can  be  assigned by specific agents called \emph{entruster agents}.

\begin{figure*}
    \includegraphics[scale=0.77]{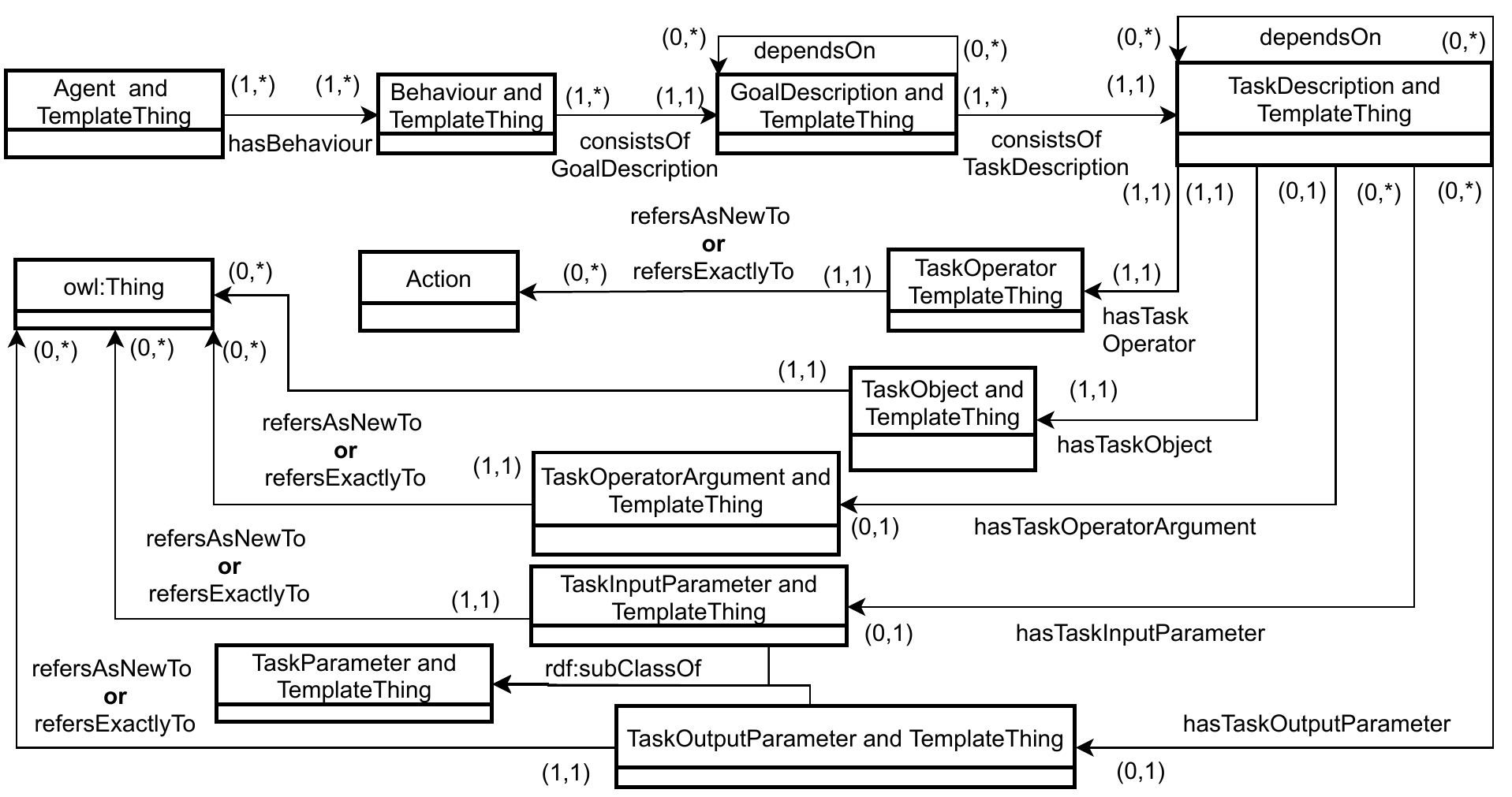}
   \caption{UML diagram of agent templates in \ONT{} 2}
    \label{fig:oasis:agent-template}
\end{figure*}

In \ONT{} 2, agent templates are defined according to the UML diagram in Fig. \ref{fig:oasis:agent-template}. 

To describe how both abstract and concrete agents perform actions, the description of agents comprises three main elements, namely \emph{behaviour}, \emph{goal}, and \emph{task}. Agent tasks, in their turn, describe atomic operations that agents may perform, including possibly input and output parameters required to accomplish them. Those elements in \ONT{} 2 are introduced by way of the following classes:

\begin{itemize} 
    \item \textit{Agent}: this class comprises all the individuals representing agents. Instances of such a class are connected with one or more instances of the class \textit{Behaviour} by means of the OWL object-property \textit{hasBehaviour}.
    
    \item \textit{Behaviour}: behaviours can be seen as collectors comprising all the goals that an agent can achieve. Instances of \textit{Behaviour} are connected with one or more instances of the class \textit{GoalDescription} by means of the object-property \textit{consistsOfGoalDescription}.
    
    \item \textit{GoalDescription}: goals represent containers embedding all the tasks that the agent can achieve. Instances of \textit{GoalDescription} comprised by a behaviour may also satisfy dependency relationships introduced by the object-property \textit{dependsOn}. Goals are connected with the tasks that form them and are represented by instances of the class \textit{TaskDescription} through the object-property \textit{consistsOfTaskDescription}.
    
    \begin{sloppypar}
    \item \textit{TaskDescription}: this class describes atomic operations that agents perform. Atomic operations are the most simple actions that agents are able to execute and, hence, they represent what agents can do within the environment. Atomic operations may depend on other atomic operations  when the object-property \textit{dependsOn} is specified. Atomic operations whose dependencies are not explicitly expressed are intended to be performed in any order. 
    \end{sloppypar}
\end{itemize}

The core of agent behaviour revolves around  the description of atomic operations represented by instances of the class \textit{TaskDescription} that characterizes the mental state corresponding to commitments. In their turn, instances of the class \textit{TaskDescription} are related with the following five elements that identify the operation: 

\begin{itemize}
    \item An instance of the class \textit{TaskOperator}, characterizing the mental state corresponding to the action to be performed. Instances of \textit{TaskOperator} are connected either by means of the object-property \textit{refersExactlyTo} or \textit{refersAsNewTo} to instances of the class \textit{Action}. The latter class describes physical actions represented by means of entity names in the form of infinite verbs (e.g., \textit{produce}, \textit{sell}). Specifically, the object-property \textit{refersExactlyTo} is used to connect the task operator with a precise action having a specific IRI,  whereas \textit{refersAsNewTo} is used to connect a task operator with an entity  representing an action of which only a general abstract   description is given (for example, an action for which only the type is known). 
    
    In the latter case, the action is also defined as an instance of the \textit{TemplateThing}: such instances are used to introduce entities that represent templates for the referred element describing the characteristics that it should satisfy. To specify the classes that the entity must be instances of, it is possible to connect the entity with instances of the desired classes using the object-property \textit{refersAsInstanceOf}.
    Finally,  \textit{TemplateThing} is a novel \ONT{} 2 class,  used to characterize all the individuals involved in the definition of  behaviour templates and to distinguish them from the entities representing concrete behaviours, plans or actions, thus eliminating the need of having separated models for those aspects.
    

    \item Possibly, an instance of the class \textit{TaskOperatorArgument}, connected by means of the object-property \textit{hasTaskOperatorArgument} and representing additional specifications or subordinate characteristics of  task operators   (e.g., \textit{on}, \textit{off}, \textit{left}, \textit{right}). Instances of \textit{TaskOperatorArgument} are referred to the operator argument by using either the object-property \textit{refersAsNewTo} or \textit{refersExactlyTo}. 
    
    \item An instance of the class \textit{TaskObject}, connected by means of the object-property \textit{hasTaskObject} and representing the template of the object recipient of the action performed by the agent (e.g., \textit{price}). Instances of \textit{TaskObject} are referred to the action recipient by specifying either the object-property \textit{refersAsNewTo} or \textit{refersExactlyTo}. 
           
    \item Input parameters and output parameters are introduced by instances of the classes \textit{TaskInputParameter} and \textit{TaskOutputParameter}, respectively. Instances of  \textit{TaskDescription} are related with instances of the classes  \textit{TaskInputParameter} and \textit{TaskOutputParameter} by means of the object-properties \textit{hasTaskInputParameter} and \textit{hasTaskOutputParameter}, respectively. Instances of \textit{TaskInputParameter} and  of \textit{TaskOutputParameter} are referred to the parameter by specifying either the object-property \textit{refersAsNewTo} or \textit{refersExactlyTo}. Moreover,  the classes \textit{TaskInputParameter} and \textit{TaskOutputParameter} are also subclasses of  \textit{TaskParameter}. 
    
\end{itemize}

\begin{sloppypar}
Finally, in the case of agent behaviour templates,  instances of \textit{Agent}, \textit{Behaviour}, \textit{GoalDescription}, \textit{TaskDescription}, \textit{TaskOperator}, \textit{TaskOperatorArgument}, \textit{TaskObject} \textit{TaskInputParameter}, and \textit{TaskOutputParameter} are also instances of \textit{TemplateThing}. 
\end{sloppypar}

\begin{figure*}
   \includegraphics[scale=0.80]{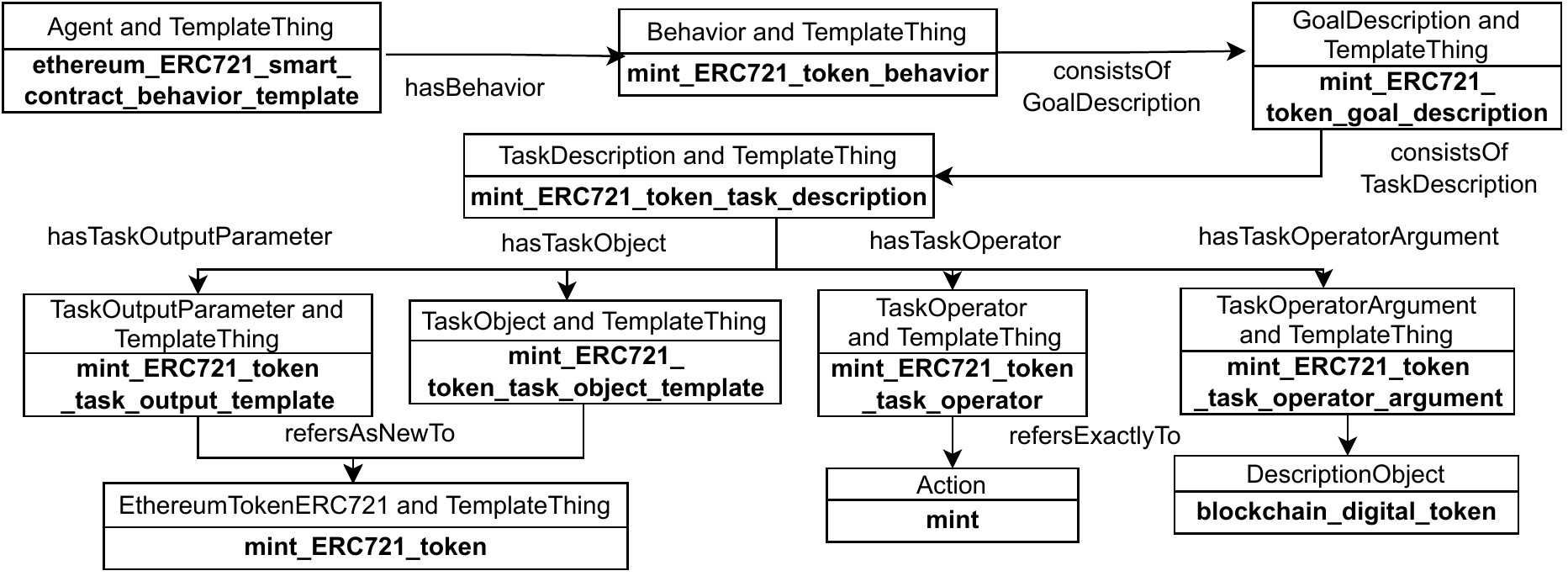}
   \caption{Example of an \ONT{} 2 agent template}
    \label{fig:oasis:agent-template-example}
\end{figure*}

Fig. \ref{fig:oasis:agent-template-example} illustrates an example of an agent template  describing the procedure of smart contracts for minting ERC-721 compliant non-fungible tokens (NFTs) on the Ethereum blockchain. The agent template described in the example comprises a single behaviour, constituted by a single goal that in its turn comprises a single task. The task, which represents the ability of the agent to mint a NFT,  provides four elements:
\begin{itemize}
\item  \textit{mint\_ERC721\_token\_task\_operator}, representing the mental state of the behaviour's action (the task operator), which is associated with the individual \textit{mint}, the latter describing the capability of generating a coin; 

\item \textit{mint\_ERC721\_token
\_task\_operator\_argument}, introducing an additional feature (the operator argument) associated with the action and represented by the individual \textit{blockchain\_digital\_token}. It describes the fact that the minting action is referred to a digital coin on the blockchain. Task operator and its argument describe together the capability of minting tokens on the blockchain; 

\item \textit{mint\_ERC721\_token\_task\_object}, representing the recipient of the operation, which is related with an instance of the class \textit{EthereumTokenERC721} by means of the object-property \textit{refersAsNewTo}. Such an instance comprises all the features that the recipient of the minting action should own, that is being a ERC-721 token. In fact, the concrete actions implementing the behaviour template for minting tokens are supposed to generate ERC-721 compliant tokens that therefore can provide additional properties, as for example the description of specific physical assets;  
\item \emph{mint\_ERC721\_token\_task\_output}, representing the output of the operation, namely the same token recipient of the \emph{mint} action.
\end{itemize}

In the second step, concrete agent behaviours are defined either by instantiating one or more templates or from scratch. In \ONT{} 2, the modeling pattern of concrete behaviours has a structure analogous to the one of behaviour templates, illustrated above, with the difference that 
individuals used to define a concrete behaviour are instances of the class 
\textit{BehaviourThing} (instead of instances of the class \textit{TemplateThing}).

\begin{figure*}
   \includegraphics[scale=0.84]{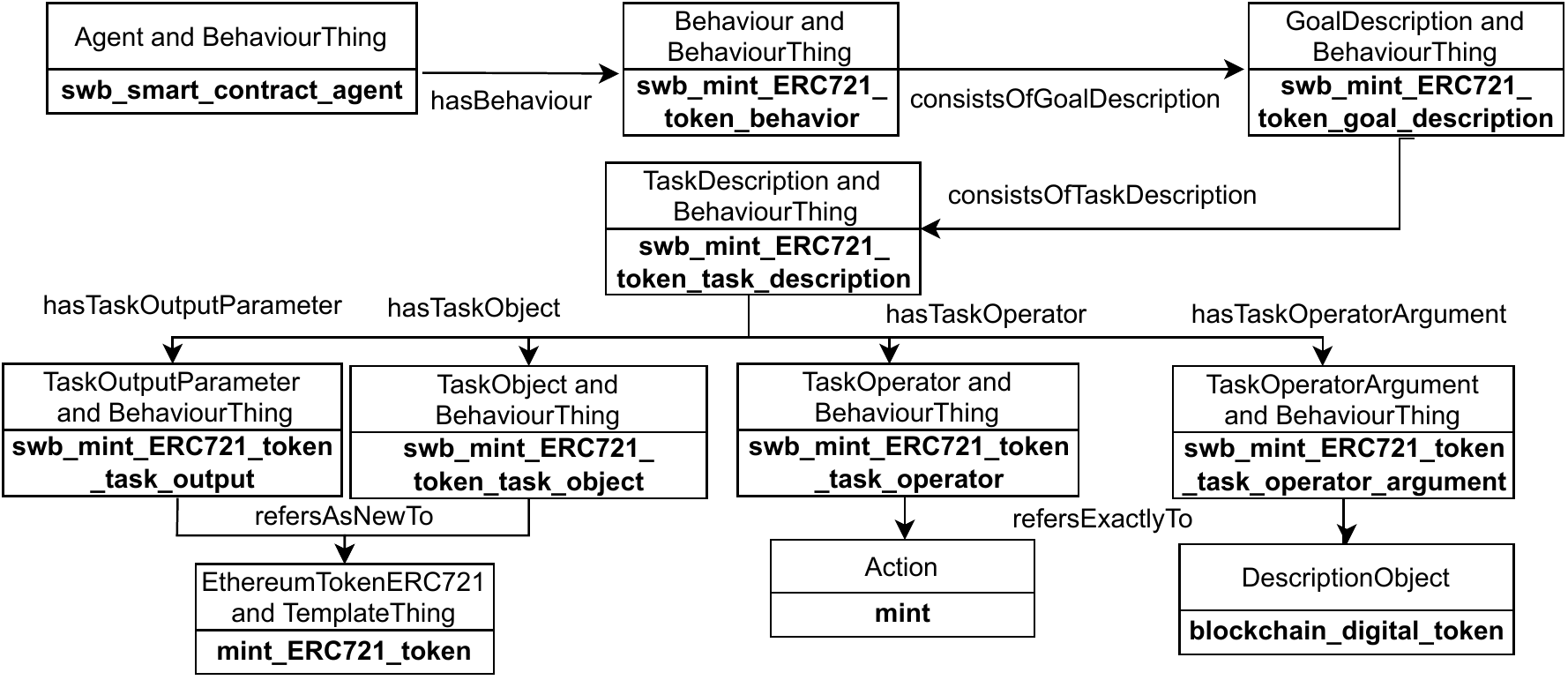}
   \caption{Example of \ONT{} 2 concrete behaviour}
    \label{fig:oasis:agent-behaviour-example}
\end{figure*}

An example of concrete behaviour is illustrated in Fig. \ref{fig:oasis:agent-behaviour-example}.\footnote{The smart contract is freely inspired from the one released by the Sicilian Wheat Bank (SWB) S.p.A.} The figure depicts a concrete behaviour for minting ERC-721 tokens,  devised from the template in Fig. \ref{fig:oasis:agent-template-example}. The template is implemented by the concrete behaviour that provides a graph similar to the template's one, where  the instances of \textit{TemplateThing} are replaced with instances of \textit{BehaviourThing}. The entity connected with the task object and the output parameter remain the same both in the template and in the concrete behaviour so as to establish that recipients of the concrete behaviours are objects of type \textit{EthereumTokenERC721} or of one of its subclasses.

Concrete behaviours can be connected with the template they are drawn from. In order to describe the fact that concrete agents inherit their behaviours from a common shared template, the  instances related with the concrete behaviours are connected with the instances of the template through the sub-properties of the object-property \textit{overloads}. The association is carried out by connecting the instances of the classes:

\begin{itemize}
    \item \textit{Behaviour}, by means of \textit{overloadsBehaviour}; 
    \item \textit{GoalDescription}, by means of \textit{overloadsGoalDescription}; 
    \item \textit{TaskDescription}, by means of \textit{overloadsTaskDescription}; 
    \item \textit{TaskObject}, by means of  \textit{overloadsTaskObject}; 
    \item \textit{TaskOperator}, by means of \textit{overloadsTaskOperator}; 
    \item \textit{TaskInputParameter}, by means of \textit{overloadsTaskInputParameter};  
    \item \textit{TaskOutputParameter}, by means of the object-property \textit{overloadsTaskOutputParameter}.
\end{itemize}

\begin{figure*}
   \includegraphics[scale=0.85]{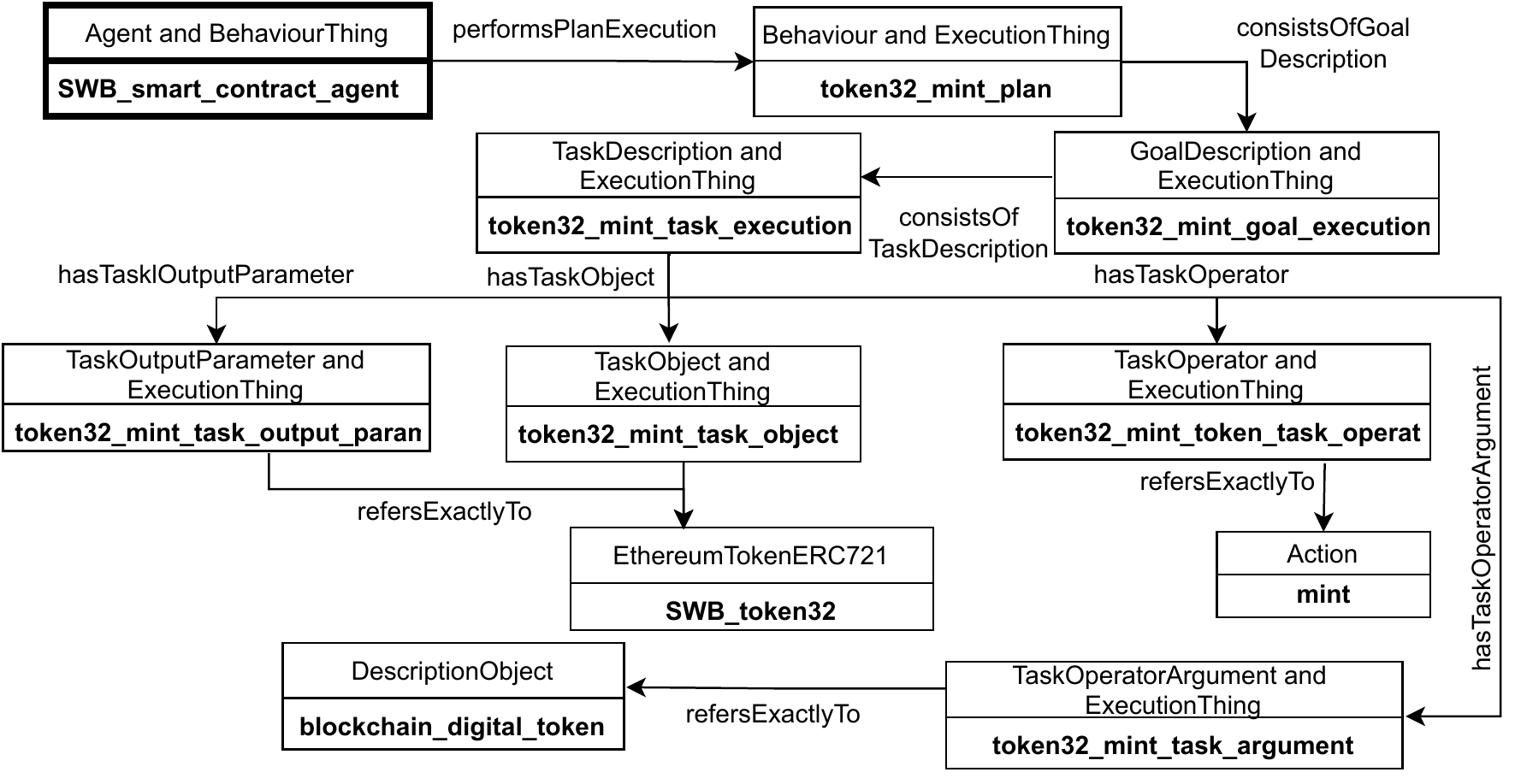}
   \caption{Example of \ONT{} 2 agent's commitment}
    \label{fig:oasis:agent-action-example}
\end{figure*}

As a last step, agent commitments devised from behaviours are introduced to describe agent actions. In \ONT{} 2, commitments are represented by adopting the same pattern presented for  abstract behaviours  with the difference that i) instances of the class \textit{TemplateThing} are instead modeled as instances of the class \textit{ExecutionThing} and ii) the agent responsible for the execution of the action is related with the plan representing the commitment by means of the object-property \textit{performsPlanExecution}, subproperty of \textit{performs}. 
The class \textit{ExecutionThing} is introduced to characterize all the entities involved in the definition of  concrete actions and to distinguish them from the ones introduced for templates, behaviours, and plans.
For instance, Fig. \ref{fig:oasis:agent-action-example} depicts the action of minting the token number 32 committed by the smart contract whose behaviour is illustrated in Fig. \ref{fig:oasis:agent-behaviour-example}. In the example, suitable instances of \textit{ExecutionThing} that represent the commitment stages take the place of the instances of \textit{BehaviourThing}, introduced in the concrete behaviour. Moreover, the template representing general Ethereum tokens has been replaced with an instance representing the token number 32, namely \textit{SWB\_token32}.

\begin{figure*}
   \includegraphics[scale=0.82]{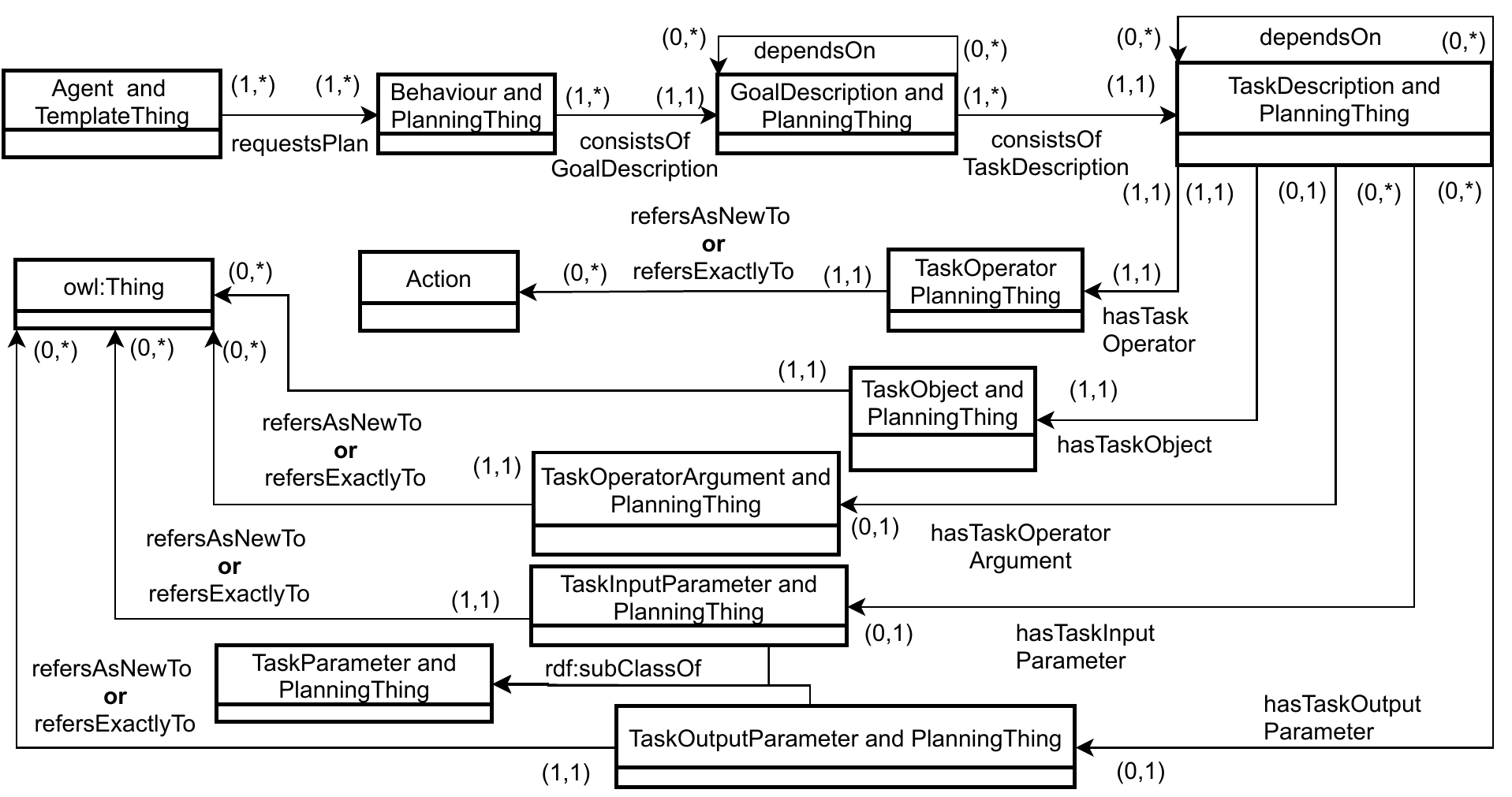}
   \caption{UML diagram of agent plans in \ONT{} 2}
    \label{fig:oasis:plan}
\end{figure*}

\begin{sloppypar}    
In order to relate agent commitments with the behaviour from which they are drawn, subproperties of the object-property \textit{drawnBy} are introduced. Specifically,  \textit{planExecutionDrawnBy} connects the instance of \textit{GoalDescription} of the agent action to its analogue of agent behaviour; much in the same way, \textit{goalExecutionDrawnBy} connects the instance of the class \textit{GoalDescription} of the commitment with its analogue, while \textit{taskExecutionDrawnBy}, \textit{taskObjectDrawnBy}, \textit{taskOperatorDrawnBy}, \textit{taskInputParameterDrawnBy}, and \textit{taskOutputParameterDrawnBy} are introduced for \textit{TaskDescription}, \textit{TaskObject},  \textit{TaskOperator}, \textit{TaskInputParameter}, and \textit{TaskOutputParameter}, respectively.
\end{sloppypar}

\begin{sloppypar}  

Agents that desire to engage peers with performing actions can submit \emph{plans}, namely descriptions of actions that they wish to be realized. For plans, a model analogous to the one describing behaviours have been introduced, where instances of \textit{BehaviourThing} are replaced with instances of \textit{PlanningThing} and agents proposing plans are connected with the instances of  \textit{PlanningThing} and \textit{Behaviour} of the plan by means of the object-property \textit{requestsPlan} (see Fig. \ref{fig:oasis:plan}). 
The class \textit{PlanningThing} is introduced to characterise all the individuals involved in the definition of  plans and to distinguish them from the ones introduced for representing templates, behaviours, and actions.
\end{sloppypar}

\begin{sloppypar}
Usually, agents proposing plans identify the behaviours responsible for their realization beforehand, in such a way as to completely describe and trace how agent intentions are realized. In this case, the entities representing the submitted plan are related with the entities describing the responsible behaviour by means of suitable subproperties of the object-property \textit{submittedTo}, relating instances of \textit{PlanningThing} with instances of \textit{BehaviourThing} as follows: a) \textit{planDescriptionSubmittedTo}, for instances of \textit{Behaviour}; b) \textit{goalDescriptionSubmittedTo}, for instances of \textit{GoalDescription}; 
c) \textit{taskDescriptionSubmittedTo}, for instances of \textit{TaskDescription},  d) \textit{taskObjectSubmittedTo}, for instances of \textit{TaskObject}; e) \textit{taskOperatorSubmittedTo}, for instances of \textit{TaskOperator}; f) \textit{taskInputParameterSubmittedTo}, for instances of \textit{TaskInputParameter}; and g) \textit{taskOutputParameterSubmittedTo}, for instances of \textit{TaskOutputParameter}.
\end{sloppypar}

In a similar way, plans are also related with the agent's action realizing them. For this purpose, the subproperties of the object-property \textit{hasExecution} are introduced, namely \textit{hasPlanExecution}, \textit{hasGoalExecution}, \textit{hasTaskExecution}, \textit{hasTaskObjectExecution}, \textit{hasTaskOperatorExecution}, \textit{hasTaskInputParameterExecution}, and \textit{hasTaskOutputParameterExecution}.

\begin{figure*}
   \includegraphics[scale=0.7]{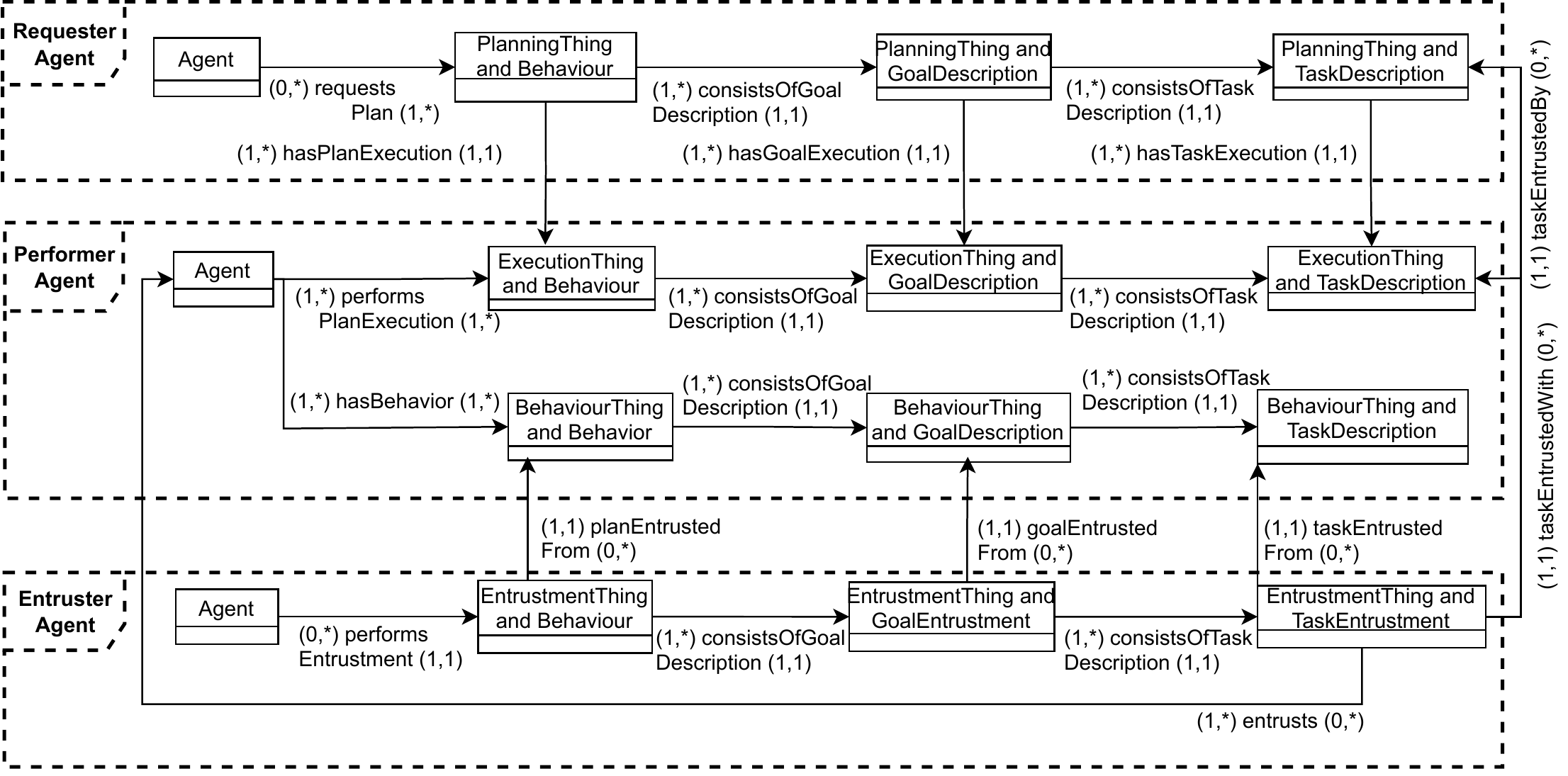}
   \caption{UML diagram of agent entrustments in \ONT{} 2}   
    \label{fig:oasis:entrustment-schema}
\end{figure*}

In \ONT{} 2, agents can assign plans to agent peers that can perform them by means of \emph{entrustments}. Entrustments represent the capability of \emph{entruster agents} to engage a \emph{performer agent} so that it executes a given plan submitted by a \emph{requester agent} (for example, a domotic assistant that activates an IoT device on behalf of a human agent). The association among the behaviours of entruster, requester, and performer are illustrated in Fig. \ref{fig:oasis:entrustment-schema}. The entrustment is introduced by means of a graph that retraces  the structure of that of agent behaviours, where instances of \textit{BehaviourThing} are replaced with instances of \textit{EntrustmentThing}. In order to associate a plan with the behaviour of the agent chosen to perform it, the elements of the entrustment are connected to a) the related elements of the plan, by means of the subproperties of \textit{entrustedBy}, and to b) the related elements of the behaviour of the performer agent, by means of the subproperties of \textit{entrustedFrom}. More precisely, \textit{entrustedBy} provides the object-properties a)
 \textit{planEntrustedBy}, to connect the entrustment to the requested plan; b) \textit{goalEntrustedBy}, to connect the entrustment's goal to the plan's goal; c) \textit{taskEntrustedBy}, to connect the entrustment's task to the plan's task; d) \textit{taskObjectEntrustedBy}, to connect the entrustment's task object to the plan's task object; e) \textit{taskOperatorEntrustedBy}, to connect the entrustment's task operator to the plan's task operator; f)  \textit{taskOperatorArgumentEntrustedBy}, to connect the entrustment's task operator argument to the plan's task operator argument; g) \textit{taskInputParameterEntrustedBy}, to connect the entrustment's task input parameter to the plan's input parameter; and h) \textit{taskOutputParameterEntrustedBy}, to connect the entrustment's task output parameter to the plan's task output parameter. On the other hand, \textit{entrustedFrom} provides analogous subproperties to connect each element of the entrustment to the related ones of the performer agents.

A single plan entrustment may be constituted by more than one goal entrustment, which in its turn may be constituted by one or more task entrustments, depending on how the requested plan is structured. In case that the requested plan is constituted by two or more tasks, each task can be entrusted to different agent behaviours, thus ensuring the representation of cooperative systems, where the effort of solving plans is distributed among participants. Finally, when the performer agent executes the entrusted plan, a plan execution is introduced as described above. As a consequence, the plan entrustment is associated with the plan execution by connecting the elements of the plan entrustment to the related elements of the plan execution through the subproperties of \textit{entrustedWith} in an analogous way to \textit{entrustedBy} and \textit{entrustedFrom}.
Additionally, the task of the plan entrustment is associated with the performer agent by means of the object-property \textit{entrusts}.



\section{\ONT{} so far} \label{sec:contrib}

The first version of \ONT{} was presented in~\cite{woa2019}, together with the architecture design and implementation of a domotic assistant based on it, called \OLDFW.\footnote{The latest version of \OLDFW\ is called \FW{} \cite{clara}. } Our approach represents a first foundational contribution to the adoption of Semantic Web technologies for defining a transparent communication protocol among agents. The proposed protocol was based on the exchange of RDF-based fragments of \ONT{}, each consisting of a description of a request to be fulfilled, by means of suitably constructed queries, against the description of the agent's behaviour selected to satisfy it. 

In~\cite{idc2021}, the \ONT{} ontology is extended with \emph{Ontological Smart Contracts} (in short, OSCs) and conditionals. OSCs are intended as agreements among agents expressed through suitable ontological fragments that permit to establish responsibilities and authorizations among agents. Conditionals have many purposes, namely they are used  a) to restrict and limit agent interactions,  b) to define activation mechanisms that trigger agent actions, and c) to define constraints and contract terms on OSCs. 

Additionally, OSCs can be secured through digital decentralized public ledgers such as the blockchain. For this purpose, \cite{idc2021} also sketches the architecture of a framework based on the \emph{Ethereum} blockchain and the \emph{Interplanetary File System}. The framework leverages a suitable Ethereum smart contract to claim and transfer  non-fungible tokens (NFT) compliant with the Ethereum ERC-721 standard, representing the ownership of OWL ontological fragments  stored on the IPFS. Moreover, smart  contracts allow one to verify unauthorized modifications both on the hierarchy of imported ontologies and on the SPARQL queries defined to verify the OSCs. The work also represents a first approach towards the definition of a \emph{Semantic Blockchain}, where operations over smart contracts are semantically represented. For this purpose, in \cite{idc2021-2}, the definition of digital contracts is extended over the blockchain to include smart contracts, intended as programs running on the blockchain and interpreted as digital agents in an \ONT{} fashion, including their operational semantics and tokens exchanged through them. It also advances the first approach to formalize ERC-721 compliant smart contracts through Semantic Web ontologies by leveraging, in turn, the behaviouristic approach pursued by \ONT{}. The main capabilities of ERC-721 compliant smart contracts are represented and exploited to fully enable a \emph{Semantic Blockchain} that provides agents and humans with means for probing the Ethereum blockchain, thus discovering smart contracts and their interactions, transactions, exchanges of cryptocurrencies and tokens, on the one hand, and for building oracles for distributed ledgers, on the other. The proposed approach is part of the  POC4COMMERCE project \cite{githubproj} funded by the NGI-ONTOCHAIN consortium \cite{ontochain}.
POC4COMMERCE, which leverages \ONT{} as foundational ontology, is presented in \cite{gecon21}, while its design and analysis can be found in \cite{pendSWJ} and in \cite{pendAO}. In the context of the POC4COMMERCE project, a semantic search engine harnessing the expressive power of \ONT{} for the e-commerce realm is under development, currently standing at TRL3 (Technology Readiness Level 3). 

Recently, an extension of \ONT{} towards \ONT{} 2 as described below has been presented in \cite{woa2022}.

\ONT{} 2 enhances \ONT{} thanks to a substantial refactoring to re-shape behaviours, plans and actions. In \ONT{}, behaviour templates, behaviours, plans, and actions are introduced by four distinct graphs rooted in the classes \textit{BehaviourTemplate}, \textit{Behaviour},  \textit{Plan} and \textit{PlanExecution}, respectively (see \cite{woa2019} for the UMLs). Instead, \ONT{} 2 introduces the novel classes \textit{TemplateThing}, \textit{BehaviourThing}, \textit{PlanningThing}, and \textit{ExecutionThing} to characterize the individuals defining behaviour templates, behaviours, plans, and actions, respectively. For the four mentioned aspects of agents, \ONT{} 2 adopts the same graph involving the classes \textit{Behaviour}, \textit{GoalDescription}, and \textit{TaskDescription}. As a consequence, the classes \textit{TaskFormalInputParameter} and \textit{TaskFormalOutputParameter} used to define parameters of behaviours in \ONT{} are replaced by the classes \textit{TaskInputParameter} and \textit{TaskOutputParameter}, adopted in combination with the class \textit{BehaviourThing}, while the classes \textit{TaskActualInputParameter} and \textit{TaskActualOutputParameter} used to define parameters of agent actions are replaced by the classes \textit{TaskInputParameter} and \textit{TaskOutputParameter}, used in combination with the class \textit{ExecutionThing}. Concerning agent templates, the input and output parameters are introduced by way of the classes \textit{TaskInputParameter} and \textit{TaskOutputParameter}, respectively, in combination with the class \textit{TemplateThing}. Finally, the object-properties connecting the agent mental states (namely, \textit{consistsOfGoalDescription}, \textit{consistsOfTaskDescription}, \textit{hasTaskInputParameter} and \textit{hasTaskOutputParameter}) have been unified, thus removing the idiosyncrasy resulting from representing strictly correlated aspects (i.e., abstract behaviour, concrete behaviour, action, and plan) through distinct models.

Additionally, the new approach proposed by \ONT{} 2 streamlines the representation of agents, also clarifying their relationships and interactions, thus bringing coherence to the model and making it more compact, smooth, and consistent. Incidentally, the novel model supports the definition of more efficient and simple queries. Most importantly, \ONT{} 2 introduces the assignment of plans and, at the same time, models the constitutional elements of blockchains, which were missing in \ONT{}.

\section{Related Work} \label{sec:related}

The integration of agent systems and the Semantic Web  has been investigated in several  contexts~\cite{Hendler2001, Hadzic2014, appOnt} and the advantages of ontology-based applications have been recognized in the  realm of MAS~\cite{Tran08}. 
Ontologies for MAS have been modeled by taking approaches similar to those of \emph{Agent-Oriented Software Engineering} (AOSE)~\cite{cossentino11}, a software engineering paradigm for the development of complex MAS based on the abstraction of agent roles and on their organizations. 
In~\cite{Freitas2017}, an ontology for agent-oriented software engineering is proposed together with a tool that uses the ontology to generate programming code for MAS. The two approaches do not examine agents and their interactions in detail and do not take into account an appropriate modeling of agent communication, which, by contrast OASIS does since its inception thanks to a protocol based on the exchange of suitable RDF fragments.

Some results attempt to bring uniformity and coherence to the increasing volume and diversity of information in a specific domain, but domain-legacy generally makes them not applicable to other contexts even if they provide interesting insights for a general approach. The downside of those models lies in the absence of relationships between agents and their commitments, which instead represents the core of agent-oriented representations. That implies the inability of finding agents/services with specific capabilities, invoking them, and enabling their interoperability, which, by contrast is smooth in \ONT{}.

The authors of~\cite{GARCIASANCHEZ2008848}  propose an ontology-based  framework for seamlessly integrating agents and Semantic Web Services, focusing on biomedical information, while  an infrastructure to allow agent-oriented platforms to access and query domain-specific OWL ontologies is presented in~\cite{freitas17b}. An approach to design scalable and flexible agent-based manufacturing systems integrating automated planning with multi-agent oriented programming for the \emph{Web of Things} (WoT) is introduced in~\cite{ciortea18}. 
Concerning the \emph{Internet of Things} (IoT), ontological approaches mainly focus on the description of sensors, with the purpose of collecting data associated with them for generating perceptions and abstractions of the observed world~\cite{garvita}. 
A comprehensive ontology for representing IoT services is presented in~\cite{wei} together with a discussion on  how it can be used to support tasks such as service discovery, testing, and dynamic composition, taking into account also parameters such as \emph{Quality of Services} (QoS), \emph{Quality of Information} (QoI), and IoT service tests. 
A unified semantic knowledge base for the Internet of Things (IoT) is proposed in~\cite{unified}, capturing the complete dynamics of IoT entities, as for instance enabling semantic searching while hiding their heterogeneity. 
%
%
%
%
All these works are valuable and have influenced the gestation of \ONT{}. In consequence, our ontology is most general, offering an extensive range of constitutional elements to represent virtually any domain that is amenable to a behaviouristic formalisation, including web services, Web of Things (WoT), IoT, and quality evaluation.

In the realm of WoT, the W3C advances a formal model and a common representation for WoT descriptions based on a small vocabulary that makes it possible both to integrate diverse devices and to allow diverse applications to interoperate~\cite{wotw3c}. The representation system provides a way to expose the state of an object and to invoke functions that, however, must be known in advance. This may complicate the task of invoking agents that would like to join the environment in a plug-and-play manner. Moreover, the provided schema does not fully allow agents to interact according to the specific roles they aim to play. This is clearly possible in OASIS.


\begin{sloppypar}
Worthy to be mentioned, conjoining the blockchain with ontologies~\cite{English2015BlockCT, CanoCimmino19} is a recent research area, mainly focused in developing a characterization of blockchain concepts and technologies through Semantic Web but without a precise analysis of the operational semantics of smart-contracts. In~\cite{Kruijff2017},  a theoretical contribution looking at the blockchain through an ontological approach has been provided. In~\cite{RSICPD18}, the authors propose a blockchain framework for SWoT contexts settled as a \emph{Service-Oriented Architecture} (SOA), where nodes can exploit smart contracts for registration, discovery, and selection of annotated services and resources.  Other works aim at representing ontologies within a blockchain context. For instance, in~\cite{KimL18} ontologies are used as a common data format for blockchain-based applications such as the proposed provenance traceability ontology, but are limited to describe implementation aspects of the blockchain. Therefore, the semantic description of smart contracts and their actions is missing from the literature, and this is one of the major results brought by \ONT{} through its formalisation of smart contracts conceived as agents modeled according to the approach pursued by the behaviouristic vision. 
\end{sloppypar}

\section{Conclusions} \label{ref:conclusions}
This paper presented the epistemological choices of the latest release, \ONT{} 2, of \emph{Ontology for Agents, Systems and Integration of Services}. \ONT{} 2 is a foundational ontology that takes the behaviouristic approach from the \emph{Theory of Agents} and inherits the related mentalistic notions to represent agents. The  paper focused on how \ONT{} 2 represents agent templates, concrete agents, agent commitments, plans, and entrustments (underlining the differences with respect to the previous version). It also introduced the representation of plan entrustments, a new feature of \ONT{} 2, and illustrated the main contributions of \ONT{} 2 and how it evolved since its first release, in particular as a consequence of its adoption in the realm of ontologies for blockchains. In fact, the NGI-ONTOCHAIN consortium aimed at delivering the ontological foundation of a blockchain-oriented e-commerce by adopting \ONT{} 2 and its behaviouristic approach.

Many advancements of \ONT{} 2 are in progress. Representing how agents reach consensus is one of the future challenges, together with the modeling of their behaviours. We shall also consider how to model in \ONT{} 2 agent roles, modular behaviours, and processes. In addition, we intend to apply agent conditionals to represent security constraints for cybersecurity threat contexts, in particular for the purpose of semantically representing authentication and confidentiality properties for agents. We shall consider how to integrate \ONT{} 2 with the main frameworks such as JADE \cite{bergenti20} to automatically generate agents and artifacts, and how it can be exploited by \emph{CArtAgO} \cite{Ricci2009}, a framework for building shared computational worlds.
\bibliographystyle{ios1}
\bibliography{biblio}

\end{document}